\documentclass[sigconf]{acmart}

\AtBeginDocument{%
  }

\usepackage{tikz}
\usetikzlibrary{positioning, arrows.meta, fit}
\usepackage{cleveref}
\usepackage{subcaption}
\usepackage{booktabs}
\usepackage{multirow}
\usepackage{makecell}

\usepackage{graphicx} 
\usepackage{amsmath}
\usepackage{geometry}
\usepackage{xcolor}
\usepackage{mathtools}
\usepackage{amsthm}
\usepackage{graphicx}
\usepackage{float} 
\usepackage{xcolor}
\usepackage{tikz}
\usetikzlibrary{decorations.pathreplacing,arrows.meta,calc}

\definecolor{peach}{RGB}{245,205,185}

\setcopyright{none}
\renewcommand\footnotetextcopyrightpermission[1]{}

\begin{document}

\title{Conditional Neural SDEs for Financial Time-Series Generation}

 \author{Sina Molavipour, Alireza M. Javid, Björn Löfdahl, Mikhail Nechaev}
 
 \email{{sina.molavipour,alireza.javid,bjorn.lofdahl,mikhail.nechaev}@seb.se}
\authornotemark[1]
 \affiliation{%
   \country{}
 }


\begin{abstract}
Conditional generation of time-series paths is a central problem in probabilistic modeling, where the goal is to simulate future trajectories that are both realistic and responsive to external context. It has direct applications in finance and economics. Neural stochastic differential equations provide a natural framework for this task by combining continuous-time stochastic dynamics with learnable drift and diffusion functions. However, directly modeling the target state can be unstable in conditional settings, especially when the target and conditioning variables have strong level dependence, limited training data, or different dimensions. We propose a residual-anchor Neural SDE architecture that decomposes the target into a learned conditional anchor and a stochastic residual. The anchor captures the context-dependent baseline, while the Neural SDE (Stochastic Differential Equation) models the residual dynamics around this baseline. This separation reduces the burden on the stochastic dynamics and generalizes spread-based representations to more general conditioning settings. 
We evaluate the approach on real-world financial time series, including conditional interest-rate scenarios and a comparison with a time-series foundation model. The results show that the proposed model can generate plausible conditional paths, maintain competitive point accuracy, and provide calibrated uncertainty estimates.
\end{abstract}


\maketitle

\pagestyle{plain}

\section{Introduction}
Scenario generation is central to many activities in financial institutions. Business units such as risk management, fixed income, credit analysis, and market research use scenario generation methods to study the possible future behavior of financial variables, including their trends, volatility, and dependence on other market factors. For example, interest rate curves are important inputs for the valuation and risk analysis of bonds, futures, and derivatives. However, realized rates can be affected by market movements, geopolitical events, and changes in monetary policy. Therefore, scenario generation models are used to describe plausible future paths and to quantify the uncertainty around them.

Classical approaches such as covariance-based Gaussian shock models are transparent and easy to implement, but they often rely on linear dependence and constant covariance structures. These assumptions can be restrictive in conditional scenario generation, especially for high-dimensional targets such as yield curves or for models with several conditioning factors.

Moreover, conventional methods such as Gaussian shock models may fail to capture nonlinear or asymmetric responses to conditioning information. For example, a positive and a negative move of the same magnitude in a conditioning factor, such as (+5) and (-5) basis points (bps), may lead to symmetric changes in the simulated target variable under a linear Gaussian model. In reality, the response can depend on the current market regime, the recent path of the variables, and the level of the target itself. A useful conditional scenario generator should therefore incorporate context when translating conditioning scenarios into target scenarios. 

Financial scenario generation has a long connection to stochastic processes. In interest rate modeling, classical continuous-time models such as the Vasicek model describe rate dynamics through mean-reverting diffusions, while extensions such as Hull-White and Heath-Jarrow-Morton models provide more flexible term structure dynamics \citep{vasicek1977,cox1985,hullwhite1990,heath1992}. These models are attractive because they are interpretable, analytically tractable in many settings, and easy to simulate. However, their parametric structure also limits the form of the drift, volatility, and response to conditioning information. Similarly, covariance-based and Gaussian shock models are useful benchmarks for risk analysis, but they often rely on linear dependence and constant covariance assumptions. As a result, they may not capture nonlinear, asymmetric, or path-dependent reactions to market scenarios.

Recent developments in neural differential equations provide a natural way to relax some of these restrictions. Neural ODEs (ordinary differential equations) replace fixed parametric dynamics with neural vector fields, and Neural SDEs (stochastic differential equations) extend this idea to stochastic dynamics by learning both the drift and diffusion functions from data \citep{chen2018,tzen2019}. This makes Neural SDEs a useful candidate for scenario generation, because they model not only the expected evolution of a financial variable but also the uncertainty around future paths. Related work has also studied neural SDEs as continuous-time generative models and developed efficient training methods for stochastic differential equations \citep{kidger2021,li2020}. In finance, Neural SDEs have mainly been explored for applications such as option pricing, calibration, and Monte Carlo simulation. Their use for conditional scenario generation of financial rates, especially under externally supplied future conditioning paths, remains less developed.

In parallel, the time-series forecasting literature has developed powerful probabilistic models for multi-horizon prediction. Models such as DeepAR and Temporal Fusion Transformers can incorporate covariates and learn predictive distributions from related time series \citep{salinas2017,lim2019}. More recent diffusion-based forecasting models, such as TimeGrad, further show that generative models can produce rich multivariate predictive distributions \citep{rasul2021}. The recent development of time-series foundation models, such as TimeGPT, TimesFM, as well as BISTRO for macroeconomic forecasting, has also shown that large pre-trained models can be adapted to forecasting tasks across different domains \citep{garza2023timegpt,das2023timesfm,koyuncu2026introducing}. These methods are highly flexible and can be effective for multi-horizon forecasting.

In this paper, we propose a residual-anchor Neural SDE framework for conditional scenario generation. The main idea is to decompose the target variable into two parts: an anchor baseline inferred from conditioning information, and a residual component that is evolved by a Neural SDE. This representation allows the model to capture the conditional level implied by the covariates while reserving the stochastic dynamics for the remaining residual variation. It also makes the approach easier to extend to multi-dimensional settings, where the target and conditioning variables may not have the same dimension or a direct one-to-one correspondence.

The rest of the paper is organized as follows. In Section~\ref{sec:background}, we review existing approaches, with particular focus on diffusion models and conditional scenario generation. In Section~\ref{sec:residual_anchor}, we present the proposed residual-anchor Neural SDE model and explain how it generalizes spread-based state representations. Section~\ref{sec:experimental_setup} describes the experimental setup and implementation details. 
In Section~\ref{sec:simulations}, we evaluate the model on real-world financial 
scenarios, and compare it with the time-series foundation model BISTRO. Finally, we conclude by summarizing the main findings and outlining possible directions for future work.

\section{Background}
\label{sec:background}

Scenario generation aims to produce plausible future paths of a time series over a given horizon. Unlike point forecasting, which focuses on a single expected trajectory, scenario generation seeks to represent a distribution of possible outcomes. This is especially important in financial applications, where the uncertainty around future paths is often as important as the central forecast.

In the unconditional setting, future scenarios are generated only from the historical behavior of the target variable. In the conditional setting, the generation also depends on additional information. Let $X_t\in \mathbb{R}^{d_x}$ denote the target financial time series and $C_t\in\mathbb{R}^{d_c}$ denote a conditioning signal. The goal is to generate future paths $X_{T+1:T+H}^{(p)}$ with $p=1,\ldots,P$, from simulation time $T$ to a horizon $H$, conditional on the observed history of $X_t$ and $C_t$, and on a given conditioning path $C_{T+1:T+H}$ for the horizon period. Then each path $p$ represents one possible
future realization. In this paper, the conditioning variables are other financial time series, such as interest rate benchmarks, swap curves, exchange rates, or macro-financial indicators.

This setting is different from standard forecasting in two ways. First, the output is not only a point estimate, but a collection of simulated paths that should reflect realistic uncertainty. Second, the generated paths should react to the conditioning information in a plausible way. For example, when the conditioning rate curve moves upward, the target rate may also move upward, but the size and direction of the response can depend on the current level, the recent trend, and the historical relationship between the two series. We briefly review the main modeling approaches related to this problem in the following.

\subsection{Covariance-based Scenario Generation}
A common approach to the generation of financial scenarios is to model changes in risk factors through a covariance matrix. In its simplest form, future shocks are sampled from a multivariate Gaussian distribution calibrated to historical changes. This type of model is transparent, easy to implement, and can preserve the contemporaneous dependence between several market variables. For example, if $X_t$ denotes a vector of financial variables, with $\Delta X_t \sim \mathcal{N}(\mu, \Sigma)$, $\mu$ and $\Sigma$ are estimated from historical data.

Such models are useful as benchmarks because they capture average volatility and cross-variable covariance. However, they usually impose linear and symmetric dependence. They also do not directly model mean reversion, state-dependent volatility, or path-dependent responses to conditioning information. In conditional settings, Gaussian covariance models can be extended by conditioning one block of variables on another, but the resulting conditional mean remains linear in the conditioning shock. This limits their ability to capture nonlinear or regime-dependent responses.

\subsection{Time-series Foundation Models}
Similar to large language models, time-series foundation models use transformer architectures to learn general patterns from large collections of sequences. In forecasting, the input is not text but historical time-series values, and the model uses the available context to predict future values. Some recent models, such as TimeGPT, TimesFM, and MOIRAI, follow this direction and provide flexible tools for probabilistic or multi-horizon forecasting.

BISTRO, the BIS Time-series Regression Oracle, is a foundation model developed for macroeconomic forecasting \citep{koyuncu2026introducing}. It builds on the MOIRAI transformer architecture and is fine-tuned on a large macroeconomic data repository maintained by the Bank for International Settlements. The data include many economic and financial series across countries, such as inflation, exchange rates, interest rates, bond yields, national accounts, and labour market indicators. BISTRO aligns mixed-frequency data on a common timeline, applies suitable transformations, and supports both unconditional and conditional scenarios. In conditional mode, future paths of selected covariates can be supplied to the model, and BISTRO generates forecasts for the target variable conditional on those paths.

\subsection{Classical SDE Models}
Classical continuous-time stochastic models for interest rates provide interpretable dynamics described by SDEs, but they rely on restrictive parametric assumptions. The models are widely used for interest rates, equity prices, and derivative pricing. 
A well-established class of SDE models is the It\^{o} diffusion,
\begin{align}
    \label{eq:Ito}
    dX_t = \mu(t,X_t)\,dt + \sigma(t,X_t)\,dW_t,
\end{align}
where $W_t$ is a Wiener process, $\mu$ is \emph{drift}, and $\sigma$ is \emph{diffusion}. The drift determines the local conditional mean evolution of the process, while the diffusion determines the local conditional volatility. Together, they define both the expected path and the uncertainty around it.
The interpretation of $X_t$ depends on the modeling choice. In interest-rate models $X_t$ may represent the level of a short rate. In equity models such as geometric Brownian motion (GBM), the state may represent the asset price or its logarithm. In yield curve applications, $X_t$ may represent a vector of rates across maturities, a spread, or a low-dimensional factor representation of the curve (i.e., principal components). In this paper, $X_t$ denotes the level of the target financial signal.

An important example is the Ornstein-Uhlenbeck (OU) process, which is widely used in interest-rate modeling and is closely related to the Vasicek short-rate model:
\begin{align}
    \label{eq:vasicek}
    dX_t = -\kappa (X_t-\bar{\mu})\,dt + \sigma\,dW_t.
\end{align}
Here $\kappa>0$ is the speed of mean reversion, $\bar{\mu}$ is the long-run mean, and $\sigma$ is a constant diffusion level. The process is
mean-reverting, so when $X_t$ diverges from $\bar{\mu}$, the drift steers it back to the mean. The OU model is therefore interpretable and stable, but it is limited by its linear drift and constant volatility.

\subsection{Neural SDE Formulation}
Neural SDEs provide a more flexible alternative by learning state-dependent drift and diffusion functions from data.
In a Neural SDE, the drift and diffusion in \eqref{eq:Ito} are parameterized by neural networks. Instead of specifying $\mu$ and $\sigma$ analytically, we learn $f_\theta(\cdot)$ and $g_\phi(\cdot)$ respectively representing the drift and diffusion:
\begin{align}
    \label{eq:NSDE}
    &dX_t = f_\theta\left(X_{t}\right)\,dt + g_\phi\left(X_{t}\right)\,dW_t,
\end{align}
where $\theta$ and $\phi$ denote trainable parameters. For conditional scenario generation, the drift and diffusion can also depend on additional conditioning information. A direct windowed conditional architecture is to provide the networks with a recent history of both the target process and the conditioning process:
\begin{align}
    \label{eq:NSDE-I}
    &dX_t = f_\theta\left(\vec{X}_{t}^{(w)}, \vec{C}_{t}^{(w)}\right)\,dt + g_\phi\left(\vec{X}_{t}^{(w)}, \vec{C}_{t}^{(w)}\right)\,dW_t.
\end{align}
Here $\vec{X}_{t}^{(w)}$ representing $X_{t-w+1:t}$ denote lookback windows of length $w$ for the target and conditioning processes. 
This direct formulation leaves the model responsible for learning the conditional baseline and the residual stochastic dynamics in a single state representation. This coupling can make recursive simulation sensitive to accumulated one-step errors, particularly for strongly co-moving target and conditioning series or limited training histories. The next section introduces a residual-anchor formulation that separates these roles.




\section{Residual-Anchor Neural SDE}
\label{sec:residual_anchor}

To motivate the residual-anchor formulation, consider first a simplified setting where the target and conditioning signals are one-dimensional and strongly co-moving. In this case, directly modeling the target level $X_t$ may be less effective than modeling a relative state. A natural choice is the spread $S_t \coloneqq X_t - C_t$.
Instead of evolving $X_t$ directly, we can model the dynamics of $S_t\in\mathbb{R}$ using a conditional Neural SDE:
\begin{align*}
    &dS_t = f_\theta\left(\vec{S}_{t}^{(w)}, \vec{C}_{t}^{(w)}\right)\,dt + g_\phi\left(\vec{S}_{t}^{(w)}, \vec{C}_{t}^{(w)}\right)\,dW_t.
\end{align*}
Once the parameters are estimated, a simulated future path $\hat{X}_{t+1:t+H}$, is reconstructed from $\hat{S}_{t+1:t+H}$ and the given $C_{t+1:t+H}$.
The spread-state formulation forces the simulated target path to remain tied to the conditioning path through an explicit reconstruction equation. This can substantially improve stability in conditional simulations, because the model does not need to relearn the dominant co-movement from scratch. 

This state representation is particularly useful when the target and conditioning series have a stable long-run relationship but the short-run spread dynamics are nonlinear. However, extending the definition of spread when $X_t$ and $C_t$ have different dimensions is not trivial. We therefore generalize the spread idea by introducing a learned conditional anchor $A_t \in \mathbb{R}^{d_x}$ and defining the residual component $R_t \in \mathbb{R}^{d_x}$ as the difference between the target state and the anchor: \vspace{-2pt}
\begin{equation*}
R_t = X_t - A_t.
\end{equation*}
In particular, the anchor for the next time step $A_{t+1}$ is computed via a learnable anchor network $a_\psi(\cdot)$:\vspace{-2pt}
\begin{equation*}
A_{t+1} \coloneqq a_\psi\left(\vec{X}_t^{(w)}, \vec{C}_{t+1}^{(w)}\right),
\end{equation*}
where $a_\psi(\cdot)$ maps historical target context and shifted conditioning inputs to the baseline level of the next target state. The shifted conditioning window $\vec{C}_{t+1}^{(w)}$ allows the anchor for the next target state to use the conditioning information available for the next step. During multi-step simulation, the target window is updated recursively using the previously simulated target values, while the conditioning window is supplied by the given or synthetically generated future conditioning path.
The residual is then evolved according to the Neural SDE:
\begin{align}
    \label{eq:NSDE-residual}
    dR_t = f_\theta(R_t,C_t)\,dt + g_\phi(R_t,C_t)\,dW_t,
\end{align}
Here, $C_t$ is included in the residual dynamics so that the drift and diffusion of the residual can still depend on the current conditioning state. Note that unlike Equation \eqref{eq:NSDE-I}, the full conditioning history of length $w$ is not passed directly into the residual SDE, which keeps the drift and diffusion networks smaller and separates the role of the anchor from the role of the stochastic residual dynamics.


To train the model on discretely observed financial time series, we discretize \eqref{eq:NSDE-residual} using Euler-Maruyama scheme:
\begin{align*}
    R_{t+1} = R_t + f_\theta\left(R_t,C_t\right)\,\Delta t + g_\phi\left(R_t,C_t\right)\sqrt{\Delta t}\,\varepsilon_{t+1},
\end{align*}
with $\varepsilon_{t+1}\sim \mathcal{N}(0,I)$. This approximation is standard for SDE-based learning from discrete observations and gives a tractable Gaussian transition likelihood as the data are observed at fixed time steps.
The model is trained using a one-step Gaussian transition likelihood. 
%
Thus the negative log-likelihood (assuming $\Delta t=1$) implies
\begin{align*}
    \mathcal{L}_{\mathrm{NLL}}(\theta,\phi,\psi)
    =
    \sum_t
        \frac{
            \left\|R_{t+1}-R_t-f_\theta\left(R_t,C_t\right)\right\|^2
        }{
            2N g_\phi\left(R_t,C_t\right)^2
        }
        +
        \log\!\left(g_\phi\left(R_t,C_t\right)^2\right),
\end{align*}
where the division and logarithm are applied componentwise for diagonal
diffusion. 

At simulation time, the model reconstructs the target as the sum of the
conditional anchor and the simulated residual. To encourage the anchor to form a useful conditional baseline, we add an anchor regularization term:
\begin{align*}
    \mathcal{L}_{\mathrm{anchor}}(\psi)
    =\frac{1}{N}\sum_{t} \left\| X_{t+1} - a_\psi\left(X_t^{(w)}, C_{t+1}^{(w)}\right)\right\|^2.
\end{align*}
So, the total loss becomes
\begin{align*}
    \mathcal{L}_{\mathrm{total}} = \mathcal{L}_{\mathrm{NLL}}(\theta,\phi,\psi) + \lambda_A \mathcal{L}_{\mathrm{anchor}}(\psi).
\end{align*}
The objective $\mathcal{L}_{\mathrm{total}}$ has an intuitive interpretation. The anchor loss encourages $a_\psi$ to accurately predict the next-step level, while the SDE likelihood models any remaining temporal dependence
and state-dependent uncertainty in the residuals. The log-variance term prevents the diffusion from becoming unnecessarily large. 
The joint objective therefore balances point-prediction accuracy with probabilistic transition modeling.

\section{Experimental Setup}
\label{sec:experimental_setup}

\subsection{Evaluation Criteria}
We evaluate the models from two complementary perspectives. First, as a sanity check for the Neural SDE formulation, we test whether the model in \eqref{eq:NSDE} can recover the dynamics of a synthetic time series generated from a Vasicek process. We analyze the estimated drift function and additionally analyze the residuals:
\begin{align}
    \label{eq:var_eps}
    \varepsilon_t \coloneqq \frac{\Delta x_t-f_\theta(x_t)\Delta t}{g_\phi(x_t)\sqrt{\Delta t}}
\end{align}
and test its distribution against Gaussian, and verify if any auto-correlation exists.  

For scenario generation, we evaluate both point-forecast accuracy and distributional behavior. For point accuracy, we compare the simulated mean path with the realized path using root mean squared error (RMSE). For distributional calibration, we compute the empirical coverage of the realized values by the interval between the 5th and 95th percentiles of the simulated paths. We also report the average width of this interval. These two quantities should be interpreted together: high coverage is not useful if it is achieved only by producing an unrealistically wide scenario cloud.

\subsection{Implementation and training}
In this section, we describe the main implementation choices used to construct the Neural SDE models. These choices apply both to the model in \eqref{eq:NSDE} and to the conditional residual-anchor model.\footnote{Code available at \url{https://github.com/s1369m/scenario_nsde}}

The drift term is modeled using a multi-layer perceptron (MLP) with \textit{SiLU} activations. This architecture is sufficient for many stationary or mean-reverting time series, such as synthetic data generated from a Vasicek process. However, when the drift contains a strong linear component, it can be useful to include this structure explicitly. We therefore use a drift head of the form:
\begin{align*}
    f_\theta (z) = \text{MLP}(z) + (\alpha+\beta^T z), 
\end{align*}
where the affine term captures linear drift behavior and the MLP captures nonlinear deviations from it.

\begin{figure}
    \centering
    \includegraphics[width=0.75\linewidth]{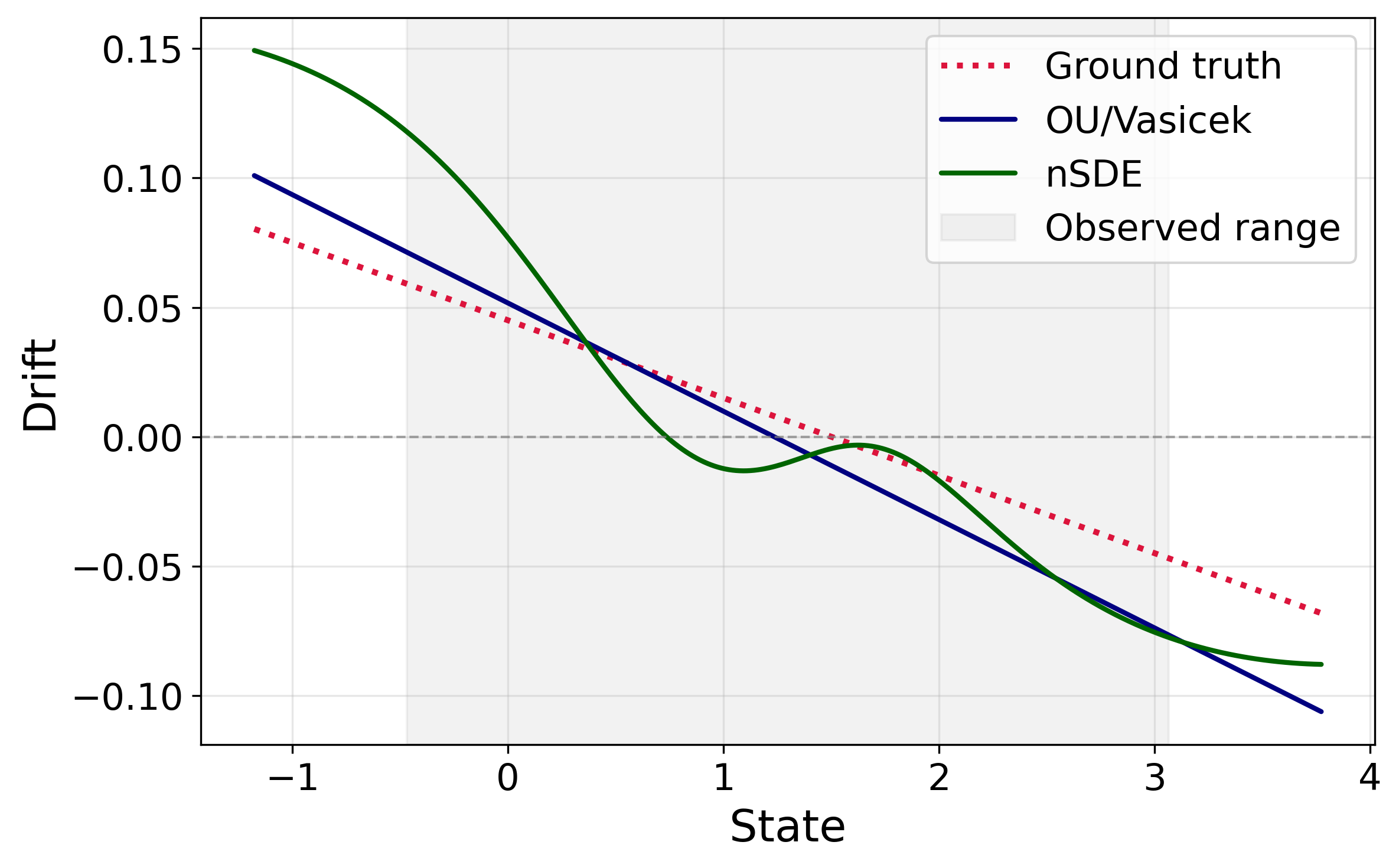}
    \caption{Estimated drift function $f_\theta(X)$, and numerical estimation of the OU model for synthetic Vasicek time-series, with respect to state $X$.}
    \label{fig:vasicek_drift_state}
    \vspace{-11pt}
\end{figure}


\begin{figure*}[t]
    \centering
    \includegraphics[width=0.7\linewidth]{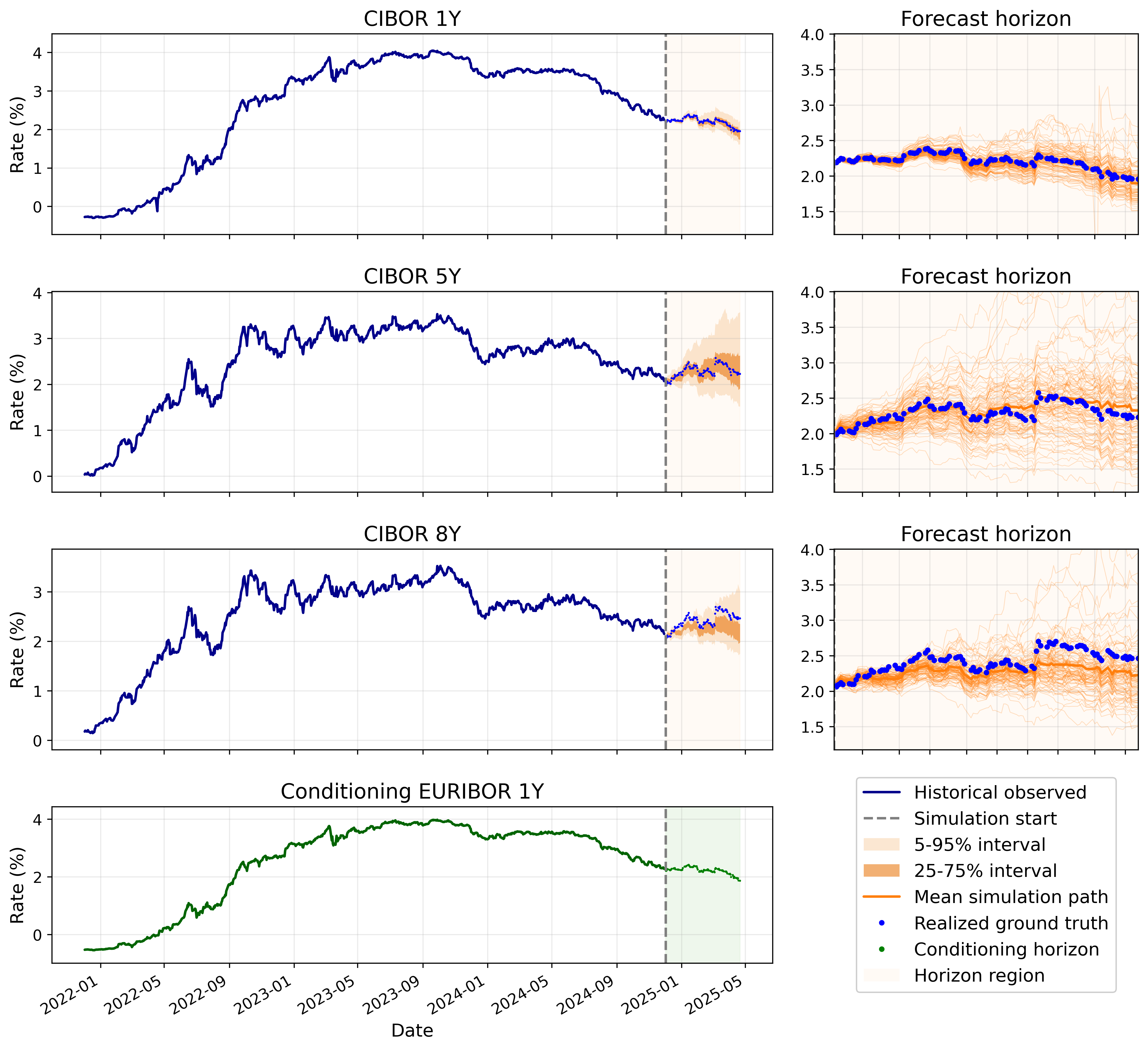}
    \caption{Backtesting conditional CIBOR scenarios generated by the residual-anchor Neural SDE. The model simulates $100$-day paths for CIBOR (1Y), (5Y), and (8Y), conditional on the realized EURIBOR (1Y).}
    \label{fig:cibor_euribor_multi}
\end{figure*}

The diffusion term must be positive and numerically stable i.e., $g_\phi(\cdot)\geq \sigma_{\min}$
for some $\sigma_{\min}>0$. Very small diffusion values can make the stochastic component negligible and may lead to unstable likelihood values. Very large diffusion values can produce overly wide scenario clouds and unstable simulated paths. To ensure positivity, we parameterize the diffusion using an MLP followed by a \textit{Softplus} transformation:
\begin{align*}
g_\phi(z)
=
\sigma_{\min}
+
\operatorname{softplus}(\text{MLP}(z)).
\end{align*}\vspace{-0.2cm}

The anchor function $a_\psi(\cdot)$ is also implemented as an MLP with SiLU activations. In our experiments, the context window length is set to $w=14$. More sophisticated architectures, such as attention-based encoders, could also be used to estimate the anchor baseline. However, the available financial time series are relatively short, and larger architectures may overfit or become unstable in this small-sample setting. For this reason, we use compact MLP architectures throughout the main experiments.

Training is performed using overlapping windows from the historical data. For each training transition, the context $\vec{X}_t^{(w)},\vec{C}_{t+1}^{(w)}$ is used to compute the next anchor.
The model parameters are then optimized by minimizing $\mathcal{L}_\mathrm{total}$ over all training windows.

In our experiments, the networks are trained using AdamW. We use 200 training epochs, a two-layer neural network with hidden size 32. Other hyperparameters, including the learning rate, hidden size, number of layers, and $\sigma_\mathrm{min}$, are selected through a grid search. Due to space limitations, we report only the selected configurations used in the final experiments. 

\section{Simulation and Results}
\label{sec:simulations}

\subsection{Synthetic Sanity Checks}
We first evaluate the Neural SDE on a synthetic time series generated from a known Vasicek process. The series contains $1000$ observations generated from \eqref{eq:vasicek} with fixed parameters $\kappa=0.03$, $\bar{\mu}=1.5$, and $\sigma=0.2$.
This experiment serves as a sanity check and as a classical benchmark. The parameters can be estimated numerically which we denote it as OU/Vasicek solver. We then train the Neural SDE on the same data and generate scenarios over a horizon of $100$ steps. Figure~\ref{fig:vasicek_drift_state} shows the value of the estimated drift function for different input states closely matches the ground-truth drift.

To evaluate the diffusion, we analyze the residuals $\varepsilon_t$ defined in \eqref{eq:var_eps}. If the fitted transition model is well calibrated, these innovations should be approximately standard normal and should not exhibit significant autocorrelation. To validate this, we employ Anderson-Darling test which is sensitive to deviations in the tails of the distibution. Larger values of Anderson-Darling statistics ($A^2$) indicate a poorer fit to the reference distribution. In this experiment, the empirical statistic is $A^2=0.411$, so we fail to reject the null hypothesis of Gaussian standardized innovations at the $5\%$ significance level.
We also apply the Ljung--Box test to check whether $\varepsilon_t$s exhibit autocorrelation across multiple lags. Using $10$ lags, the observed test statistic is $Q=12.79$. At the $5\%$ significance level, we fail to reject the null hypothesis of no autocorrelation. These diagnostics suggest that the fitted Neural SDE provides a reasonable one-step stochastic representation of the dynamics.

\subsection{Conditional Scenario Generation}
We consider several conditional scenario-generation experiments. The aim is to evaluate whether the Neural SDE can generate plausible future paths for a target financial series when a conditioning signal is supplied. 

The empirical experiments use several reference rates. CIBOR denotes the Copenhagen Interbank Offered Rate, the Danish interbank benchmark, and STIBOR is the corresponding Stockholm Interbank Offered Rate. EURIBOR denotes the Euro Interbank Offered Rate, a euro-area money-market benchmark. These rates are commonly used as reference rates in floating-rate instruments, money-market products, and interest-rate derivatives. They are also strongly related across markets, which makes them useful for testing conditional scenario-generation methods.

In the historical experiments, one rate series is treated as the target variable and another related market series is treated as the conditioning signal. For example, we study CIBOR scenarios conditioned on EURIBOR to test whether the model can generate plausible paths for a local benchmark rate given movements in a broader European rate benchmark. We also investigate whether adding the EUR/SEK exchange rate improves backtesting performance when Swedish market variables are involved. Finally, we compare the proposed method with BISTRO, a time-series foundation model proposed in \cite{koyuncu2026introducing}, using historical backtests over a ten-year period.


\begin{figure}[t]
    \centering
    \includegraphics[width=0.8\linewidth]{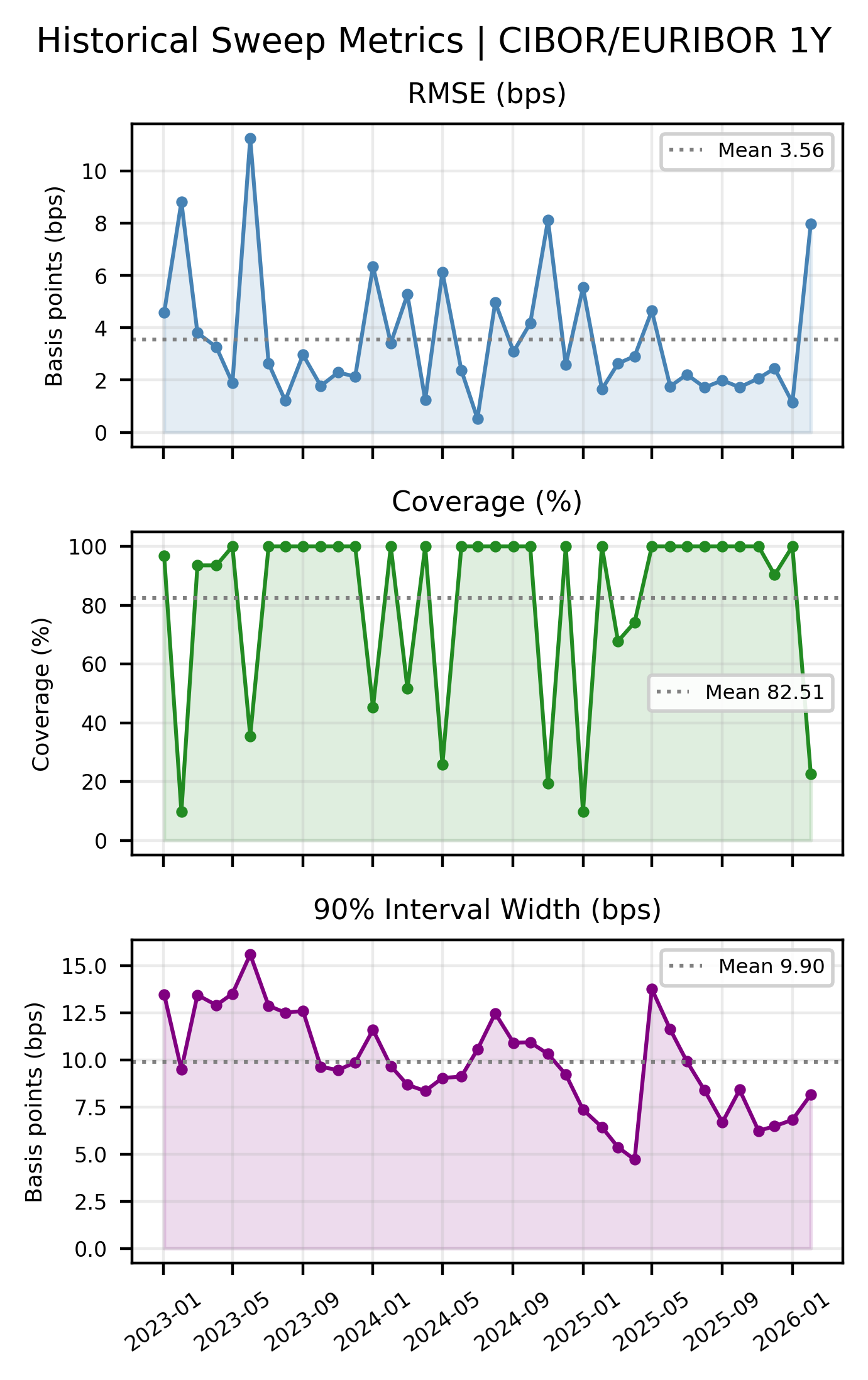}
    \caption{Monthly rolling backtest of the residual-anchor Neural SDE for CIBOR (1Y) conditioned on EURIBOR (1Y) over 2023--2026.}
    \label{fig:historical_sweep}
\end{figure}

\subsubsection{CIBOR vs. EURIBOR}

We next evaluate the residual-anchor Neural SDE on historical interest-rate data. In this experiment, the target series is CIBOR and the conditioning series is EURIBOR. These rates are economically related and often co-move, while their difference reflects local market conditions, liquidity, credit risk, and currency-specific factors. Therefore, EURIBOR can provide useful conditioning information for generating CIBOR scenarios.

One advantage of the residual-anchor formulation is that it can be extended naturally to multivariate target scenario generation. This is important for yield-curve applications, where the target object consists of rates observed across several maturities. In Figure~\ref{fig:cibor_euribor_multi}, the Neural SDE model is trained on historical data from 2022 to 2025. The target is the vector of CIBOR rates at maturities (1Y), (5Y), and (8Y), while the conditioning signal is the EURIBOR (1Y) rate. The model is then used to simulate CIBOR paths over a $100$-day horizon, conditional on the realized EURIBOR (1Y) path over the same horizon.

The results show that the simulated paths track the realized CIBOR (1Y) series more closely than the longer-maturity CIBOR series. This is expected, because the conditioning signal is also a (1Y) rate and is therefore more directly related to the (1Y) CIBOR target. For the (5Y) and (8Y) targets, the relationship with EURIBOR (1Y) is weaker and less direct. This is reflected in wider simulated scenario bands and larger deviations from the realized paths.

To provide a more comprehensive historical evaluation, we perform a rolling monthly backtest over the period 2023--2026 with CIBOR (1Y) as the target conditioning on EURIBOR (1Y). At each backtest date, the model is trained using the previous two years of historical data and then used to generate scenarios over a $30$-day horizon. Figure~\ref{fig:historical_sweep} summarizes the backtesting results. Across the backtest dates, the simulated mean paths remain close to the realized CIBOR paths, with an average RMSE of $3.56$ bps. At the same time, the uncertainty bands remain relatively narrow, with an average $5\%$--$95\%$ interval width of $9.9$ bps. These results suggest that the residual-anchor Neural SDE can produce accurate and reasonably calibrated conditional scenarios in this CIBOR--EURIBOR setting.

\subsubsection{Scenario Generation for Wider Scopes}
To test the model in a setting with richer conditioning information, we simulate STIBOR scenarios conditional on EURIBOR and the EUR/SEK exchange rate. Although STIBOR and EURIBOR are related through broader European rate-market movements, the spread between them can also be affected by country-specific factors, monetary-policy expectations, and exchange rate movements. Therefore, EUR/SEK may provide additional information about Swedish market conditions relative to the euro area.

\begin{figure*}[t]
    \centering
    \includegraphics[width=0.7\linewidth]{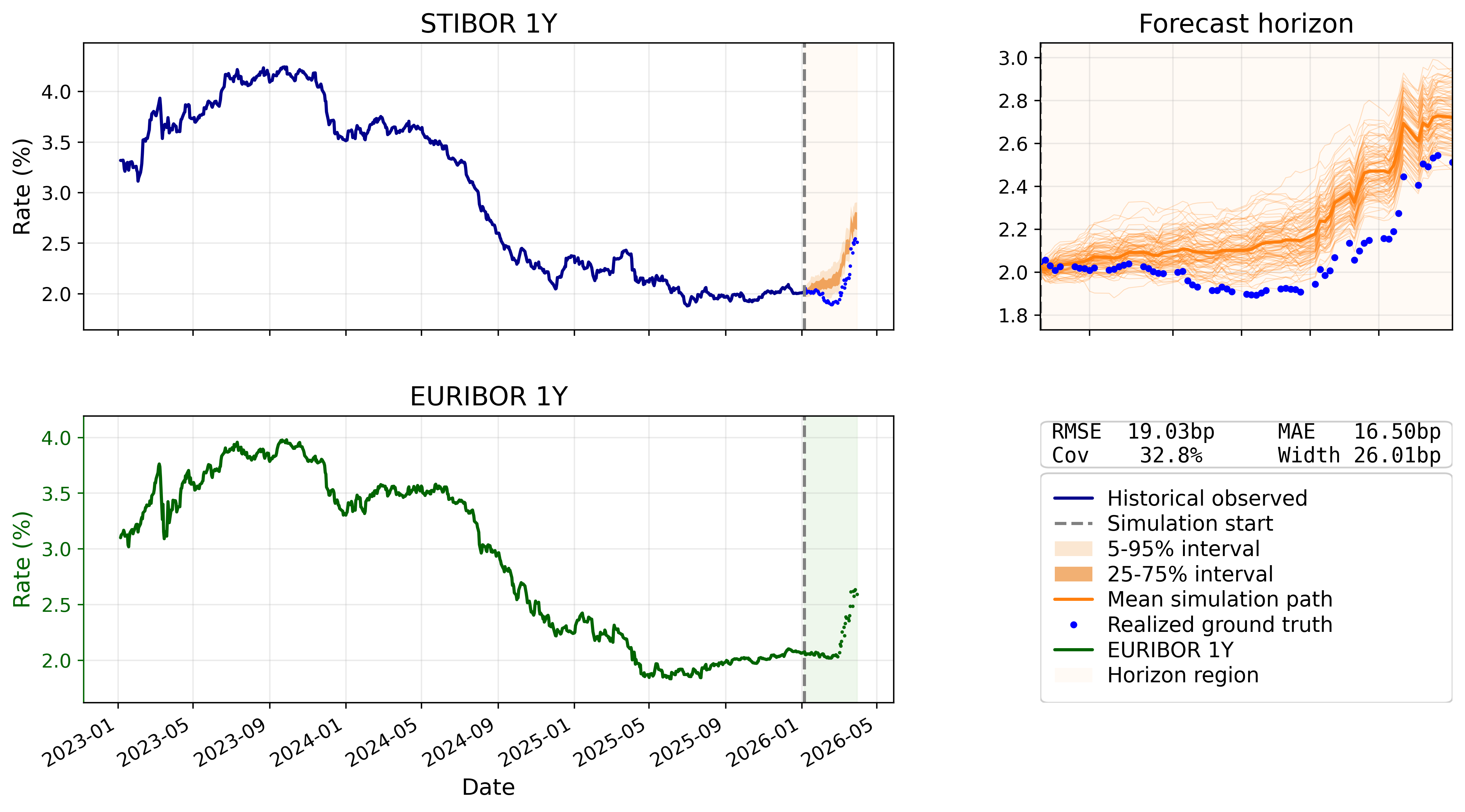}
    \caption{Conditional STIBOR (1Y) scenario generation using EURIBOR (1Y) as the conditioning signal.}
    \label{fig:STIBOR_EURIBOR}
\end{figure*}

\begin{figure*}[t]
    \centering
    \includegraphics[width=0.7\linewidth]{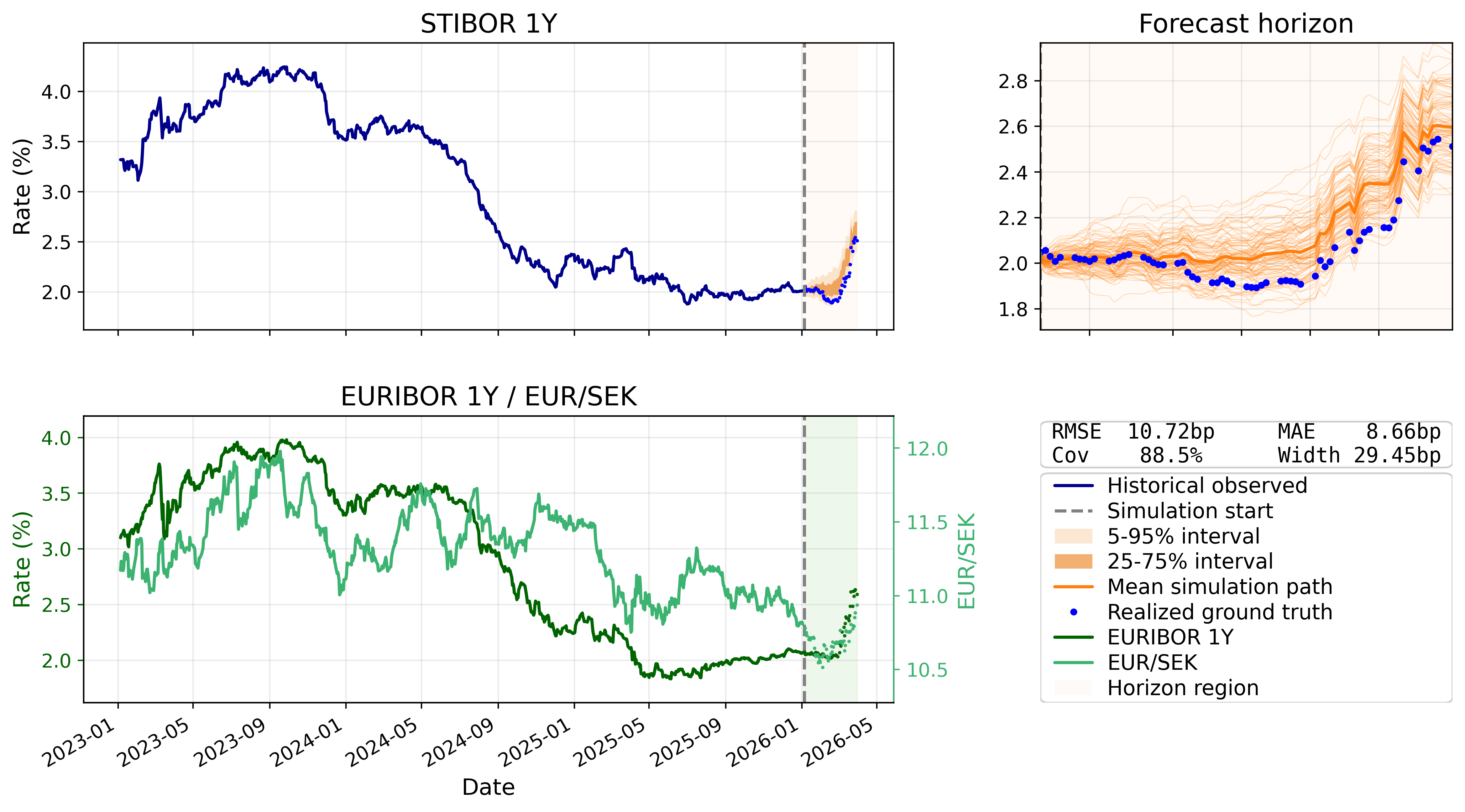}
    \caption{Conditional STIBOR (1Y) scenario generation using both EURIBOR (1Y) and EUR/SEK as conditioning signals.}
    \label{fig:STIBOR_EURIBOR_EUR}
\end{figure*}

We train the model using three years of historical data and generate STIBOR scenarios over a $60$-day horizon in 2025. We consider two conditioning setups. In the first setup, shown in Figure~\ref{fig:STIBOR_EURIBOR}, the model is conditioned only on the EURIBOR (1Y) rate. In the second setup, shown in Figure~\ref{fig:STIBOR_EURIBOR_EUR}, we add EUR/SEK as an additional conditioning signal. This allows us to test whether the residual-anchor Neural SDE can use extra conditioning information beyond the interest rate benchmark alone. In this backtest, adding EUR/SEK improves the point accuracy of the simulated STIBOR paths, as reflected by the lower RMSE.

\begin{figure*}[t]
    \centering
    \includegraphics[width=0.7\linewidth]{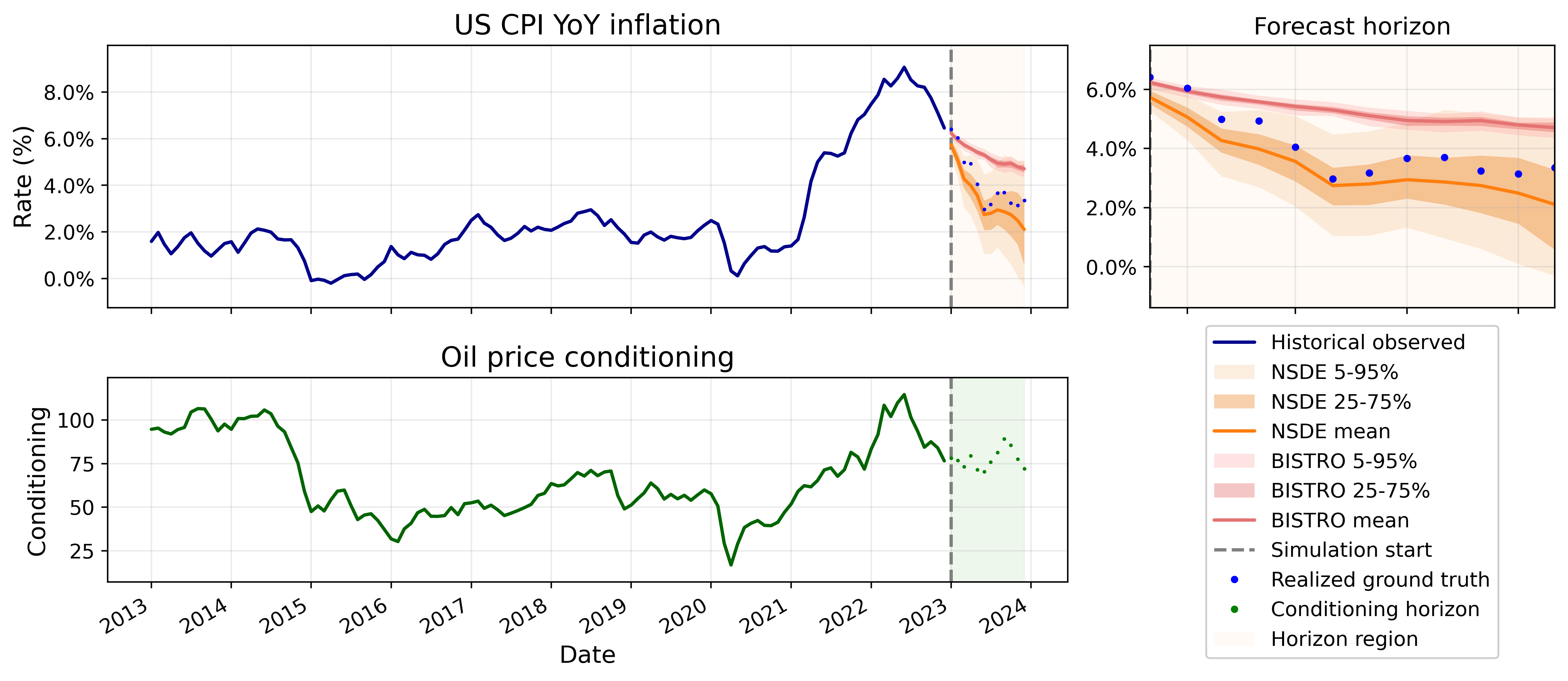}
    \caption{Comparison between our proposed Neural SDE and the BISTRO foundation model for US CPI YoY inflation scenarios conditional on the oil price path.}
    \label{fig:BISTRO_vs_NSDE}
\end{figure*}

\begin{figure}[t]
    \centering
    \includegraphics[width=0.75\linewidth]{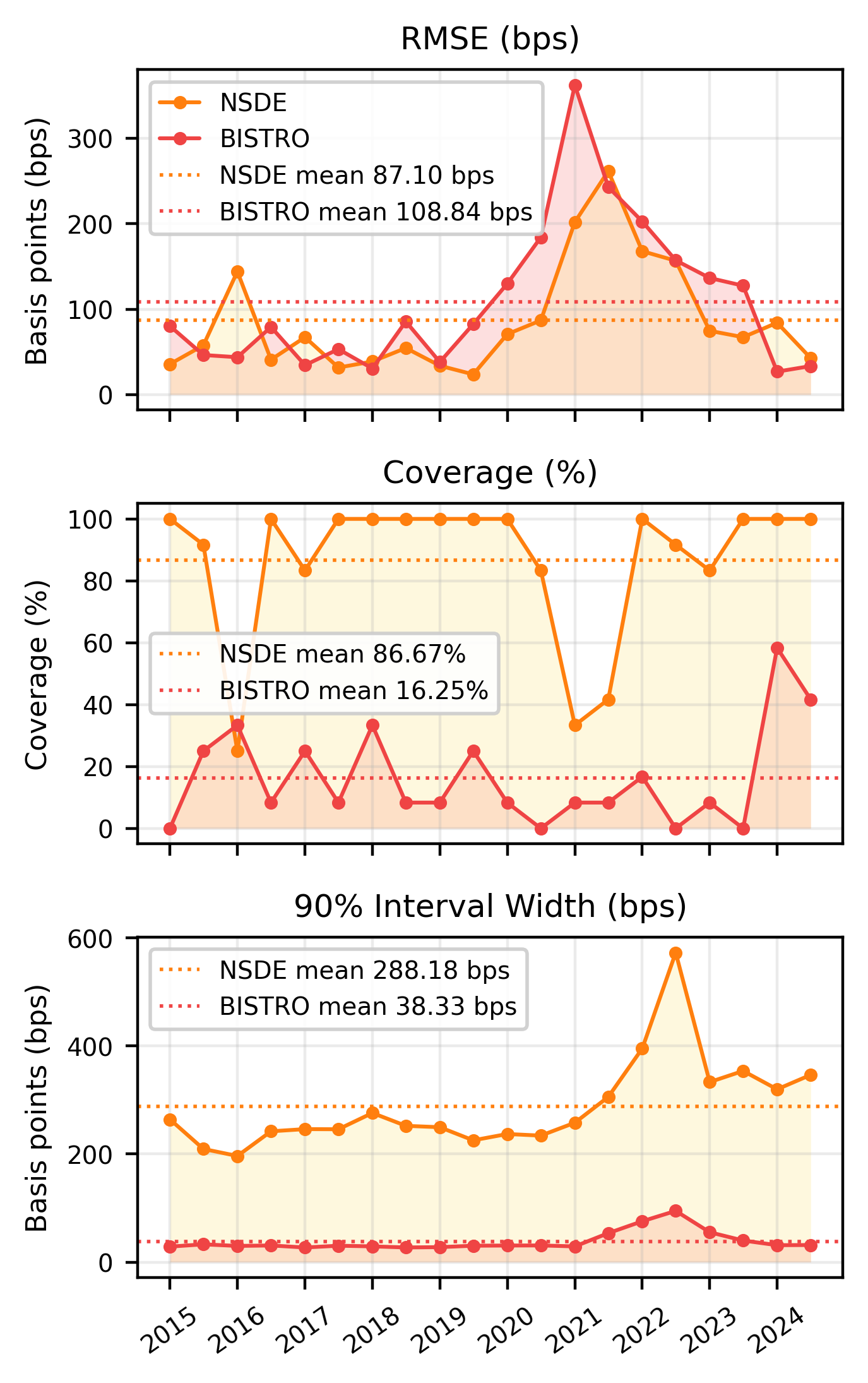}
    \caption{Historical backtest comparison between the residual-anchor Neural SDE and BISTRO for US CPI YoY inflation conditioned on oil prices. Metrics report point accuracy, coverage, and interval width.}
    \label{fig:BISTRO_vs_NSDE_sweep}
\end{figure}

Recent work on time-series foundation models has shown that large pre-trained models can be adapted to a wide range of forecasting tasks. As an additional benchmark, we compare the proposed Neural SDE approach with BISTRO, a foundation model proposed by the Bank for International Settlements for unconditional and conditional macroeconomic forecasting \cite{koyuncu2026introducing}.

For this comparison, the target variable is monthly US year-on-year (YoY) CPI inflation, and the conditioning variable is the monthly oil price. For the Neural SDE, we use a training period of $20$ years, corresponding to $240$ monthly observations, and set the context window to $w=12$ months. The forecast horizon is $12$ months. For BISTRO, we use the same context length and horizon.

Figure~\ref{fig:BISTRO_vs_NSDE} compares the two methods for forecasting CPI inflation over 2023--2024. The Neural SDE scenarios are more closely aligned with the realized CPI path in this example. We further evaluate the comparison through a historical backtest from 2015 to 2025, shown in Figure~\ref{fig:BISTRO_vs_NSDE_sweep}. Across this backtest, the Neural SDE achieves lower RMSE than BISTRO in most evaluation windows. The Neural SDE also provides higher empirical coverage, although this comes with wider scenario bands. This indicates that the Neural SDE not only tracks the realized inflation path more closely on average, but also assigns substantially more probability mass around the realized outcomes while the uncertainty bands are wider. This highlights the trade-off between point accuracy and distributional calibration.

It is important to interpret this comparison in context. BISTRO is a large foundation model trained on a broad collection of macroeconomic and financial time series, while the Neural SDE is trained only on the $240$ monthly observations available in each local training window. The comparison therefore does not suggest that the two models use the same amount of information. Rather, it shows that our Neural SDE model can provide competitive conditional scenario paths in this particular macroeconomic forecasting setup.

\section{Conclusion}
\label{sec:conclusion}

This paper introduced a residual-anchor Neural SDE for conditional financial scenario generation where the model first learns a conditioning-based anchor and then applies the SDE to the remaining residual. By isolating the conditional level baseline, the Neural SDE network only needs to learn stationary zero-mean fluctuations $Z_t$, drastically simplifying the optimization landscape over short financial sample regimes.


In historical interest-rate experiments, the model generates CIBOR scenarios conditional on EURIBOR and extends naturally to multi-maturity targets. The rolling CIBOR vs. EURIBOR backtest shows that the simulated paths remain close to realized values while maintaining relatively narrow uncertainty bands. The STIBOR experiment further shows that adding EUR/SEK as an additional conditioning signal can improve point accuracy. Finally, in the comparison with BISTRO on US CPI inflation conditioned on oil prices, the Neural SDE achieves lower RMSE in most backtest windows and substantially higher empirical coverage, although this comes with wider scenario intervals.

Overall, the results suggest that the residual-anchor representation is a useful way to combine conditional information with stochastic path simulation. The anchor captures the conditional level implied by the supplied market scenario, while the residual SDE models the remaining uncertainty around that level. This makes the method especially suitable for financial settings where interpretability and conditional responsiveness are paramount.

There are several directions for future work. First, the current experiments assume that the future conditioning path is supplied externally. A complete scenario engine would also require joint generation of the conditioning variables. Second, the current implementation uses compact MLP architectures because the available financial histories are relatively short. More expressive encoders, such as attention-based models, could be explored when longer histories or richer cross-sectional data are available. Finally, the framework can be extended to higher-dimensional curves and other financial applications where conditional path generation is needed.

\bibliographystyle{ACM-Reference-Format}
\bibliography{references}

\end{document}